\pdfoutput=1

\documentclass[11pt]{article}

\usepackage[]{ACL2023}

\usepackage{times}
\usepackage{latexsym}

\usepackage[T1]{fontenc}

\usepackage[utf8]{inputenc}

\usepackage{microtype}

\usepackage{inconsolata}

\usepackage{booktabs}
\usepackage{multirow}
\usepackage{graphicx}
\usepackage{subcaption}

\title{The Impact of Debiasing on the Performance of Language Models in Downstream Tasks is Underestimated}

\author{Masahiro Kaneko$^{1,2}$ \quad
        Danushka Bollegala$^{3,4}$\Thanks{Danushka Bollegala holds concurrent appointments as a Professor at University of Liverpool and as an Amazon Scholar. This paper describes work performed at the University of Liverpool and is not associated with Amazon.} \quad
        Naoaki Okazaki$^{2}$ \\
        $^1$MBZUAI \quad
        $^2$Tokyo Institute of Technology \quad
        $^3$University of Liverpool \quad
        $^4$Amazon \\
        {\tt Masahiro.Kaneko@mbzuai.ac.ae} \\
        {\tt danushka@liverpool.ac.uk} \quad
        {\tt okazaki@c.titech.ac.jp}
}

\begin{document}
\maketitle
\begin{abstract}
Pre-trained language models trained on large-scale data have learned serious levels of social biases. 
Consequently, various methods have been proposed to debias pre-trained models.
Debiasing methods need to mitigate only discriminatory bias information from the pre-trained models, while retaining information that is useful for the downstream tasks.
In previous research, whether useful information is retained has been confirmed by the performance of downstream tasks in debiased pre-trained models.
On the other hand, it is not clear whether these benchmarks consist of data pertaining to social biases and are appropriate for investigating the impact of debiasing.
For example in gender-related social biases, data containing female words (e.g. \textit{``she, female, woman''}), male words (e.g. \textit{``he, male, man''}), and stereotypical words (e.g. \textit{``nurse, doctor, professor''}) are considered to be the most affected by debiasing.
If there is not much data containing these words in a benchmark dataset for a target task, there is the possibility of erroneously evaluating the effects of debiasing.
In this study, we compare the impact of debiasing on performance across multiple downstream tasks using a wide-range of benchmark datasets that containing female, male, and stereotypical words.
Experiments show that the effects of debiasing are consistently \emph{underestimated} across all tasks.
Moreover, the effects of debiasing could be reliably evaluated by separately considering instances containing female, male, and stereotypical words than all of the instances in a benchmark dataset.
\end{abstract}

\section{Introduction}

Unfortunately, Pre-trained Language Models (PLMs) such as BERT~\cite{devlin-etal-2019-bert} and RoBERTa~\cite{liu2019roberta} easily learn discriminatory social biases expressed in human-written texts in massive datasets~\cite{kurita-etal-2019-measuring,zhou-etal-2022-sense,kaneko-etal-2022-gender}.
For example, if a model is given ``\textit{[MASK] is a nurse.}'' as the input, a gender biased PLM would predict ``\textit{She}'' with a higer likelihood score than for ``\textit{He}'' when filling the [MASK].
Various debiasing methods have been proposed to mitigate social biases in PLMs.
\citet{zhao-etal-2019-gender,webster2020measuring} proposed a debiasing method by swapping the gender of female and male words in the training data.
\citet{kaneko-bollegala-2021-debiasing} proposed a method for debiasing by orthogonalising the vectors representing gender information with the hidden layer of a language model given a sentence containing a stereotypical word.
\citet{webster2020measuring} showed that dropout regularization can reduce overfitting to gender information, thereby can be used for debiasing PLMs.

\begin{table}[!t]
\centering
\small
\begin{tabular}{lrrrr}
\toprule
& All & Female & Male & Occ. \\
\midrule
CoLA & 1,043 & 174 & 722 & 96 \\
MNLI & 9,832 & 3,467 & 8,875 & 1,415 \\
MRPC & 408 & 101 & 391 & 96 \\
QNLI & 5,463 & 2,149 & 5,371 & 1,066 \\
QQP & 40,430 & 7,415 & 29,638 & 3,331 \\
RTE & 277 & 113 & 269 & 94 \\
SST-2 & 872 & 187 & 691 & 75 \\
STS-B & 1,500 & 513 & 1,277 & 151 \\
WNLI & 71 & 27 & 71 & 6 \\
\bottomrule
\end{tabular}
\caption{The total number of instances containing female, male, and occupational (Occ.) words in the GLUE development data.}
\label{tbl:statistics}
\end{table}

The debiasing method should mitigate only discriminatory information, while pre-trained useful information should be retained in the model.
Evaluations in downstream tasks often employ the GLEU benchmark~\cite{wang-etal-2018-glue}, which measures the ability to understand language~\cite{kaneko-bollegala-2021-debiasing,guo-etal-2022-auto,meade-etal-2022-empirical}.
The data for downstream tasks are not selected in terms of whether they reflect the impact of debiasing.
To mitigate gender bias, data containing female words such as \textit{``she''} and \textit{``woman''}, male words such as \textit{``he''} and \textit{``man''}, and stereotypical words such as \textit{``doctor''} and \textit{``nurse''} would be most affected by debiasing.

\autoref{tbl:statistics} shows the total number of instances containing female, male, and occupational (Occ.) words in the development data in the GLUE benchmark suite~\cite{GLUE}, which is widely recognised as a standard evaluation benchmark for LLMs.
Occupational words have been used for probing LLMs for stereotypical social biases~\cite{bolukbasi2016man}.
From \autoref{tbl:statistics}, we see that the GLEU benchmark has little data related to females and occupations.
Therefore, the impact of debiasing on data related to females and occupations may be potentially underestimated when LLMs are evaluated on GLUE.

We first extract instances containing female words, data containing male words, and data containing stereotypical words from the benchmarks.
We then calculated the performance difference between the original model and the debiased model for each category and compared it to the performance difference using the entire benchmark.
The results showed that the debiased model performed worse than the original model on data related to females and occupations compared to the original model when evaluated on the entire dataset.
Therefore, existing evaluations underestimate the impact of debiasing on the performance of the downstream task.

It is important to be able to compare how well the effects of debiasing are captured in the datasets related to females, males, and occupations.
We propose a method to control the degree of debiasing of PLMs and investigate whether the performance difference between original and debiased models widens as the degree of debiasing increases.
Experimental results showed that the proportion of female, male and occupational words in the dataset is related to the susceptibility of the dataset to debiasing.

\section{Experiments}

\subsection{Debiasing Methods}

We use the following three commonly used debiasing methods in our experiments.
We apply these debiasing methods during fine-tuning in downstream tasks.
\paragraph{Counterfactual Data Augmentation (CDA) debiasing:}
CDA debiasing~\cite{webster2020measuring} swaps the gender of gender words in the training data.
For example, \textit{``She is a nurse''} is swapped to \textit{``He is a nurse''}, and the swapped version is appended to the training dataset.
This enables to learn a less biased model because the frequency of female and male words will be the same in the augmented dataset.

\paragraph{Dropout debiasing:}
\citet{webster2020measuring} introduced dropout regularisation as a method to mitigate biases.
They enhanced the dropout parameters for the attention weights and hidden activations of PLMs.
Their research demonstrated that intensified dropout regularisation diminishes gender bias in these PLMs.
They showed that dropout interfers with the attention mechanism in PLMs, and prevents undesirable associations between words. 
However, it is also possible that the model may no longer be able to learn desirable associations.

\paragraph{Context debiasing:}
\citet{kaneko-bollegala-2021-debiasing} proposed a method to debias MLMs through fine-tuning.
It preserves semantic information while removing gender-related biases using orthogonal projections at token- or sentence-level.
This method targets male and female words and occupational words in the text for debiasing.
This method can be applied various MLMs, independent of the model architectures and pre-training methods.
Token-level debiasing across all layers produces the best performance.

\begin{table*}[!t]
\centering
\small
\begin{tabular}{lrrrrrrrrrrrr}
\toprule
& \multicolumn{4}{c}{CDA} & \multicolumn{4}{c}{Dropout} & \multicolumn{4}{c}{Context} \\
\cmidrule(r){2-5}
\cmidrule(r){6-9}
\cmidrule(r){10-13}
& All & Female & Male & Occ. & All & Female & Male & Occ. & All & Female & Male & Occ. \\
\midrule
CoLA & -1.36 & \textbf{-3.42} & -2.01 & -1.45 & 0.42 & -0.14 & \textbf{-0.21} & -0.07 & -0.32 & \textbf{-0.86} & -0.71 & -0.55 \\
MNLI & -0.55 & \textbf{-0.90} & -0.71 & -0.63 & 0.23 & 0.13 & \textbf{0.01} & 0.05 & -0.05 & \textbf{-0.47} & -0.43 & -0.32 \\
MRPC & -0.96 & -1.28 & \textbf{-1.31} & -1.03 & -0.82 & \textbf{-1.12} & -1.02 & -1.04 & -0.88 & -1.01 & \textbf{-1.06} & -0.92 \\
QNLI & -1.13 & \textbf{-1.42} & -1.19 & -1.27 & -1.01 & -1.11 & -1.07 & \textbf{-1.21} & 0.25 & \textbf{-0.19} & -0.06 & -0.04 \\
QQP & -0.21 & \textbf{-0.69} & -0.32 & -0.25 & 0.53 & \textbf{0.13} & 0.47 & 0.30 & 0.14 & \textbf{-0.12} & 0.03 & -0.05 \\
RTE & -1.16 & \textbf{-1.21} & -1.02 & -1.13 & -1.01 & \textbf{-1.24} & -0.96 & -1.13 & -0.43 & -0.65 & -0.51 & \textbf{-0.73} \\
SST-2 & -0.11 & \textbf{-0.81} & -0.34 & -0.25 & 0.45 & 0.20 & \textbf{0.12} & 0.23 & 0.22 & \textbf{-0.15} & -0.02 & -0.12 \\
STS-B & -1.01 & \textbf{-1.95} & -1.34 & -1.10 & 0.21 & 0.09 & -0.03 & \textbf{-0.11} & -0.08 & -0.31 & \textbf{-0.38} & -0.34 \\
WNLI & -2.82 & \textbf{-3.07} & -2.82 & -2.71 & -2.01 & -2.21 & -2.01 & \textbf{-2.33} & -1.52 & \textbf{-1.88} & -1.52 & -1.61 \\
\bottomrule
\end{tabular}
\caption{Performance difference between the original model and debiased model for each dataset. \textbf{Bolded} values indicate the largest drop in performance of the debiased model.}
\label{tbl:diff}
\end{table*}

\subsection{Settings}

Although we use BERT (bert-base-cased\footnote{\url{https://huggingface.co/bert-base-cased}})~\cite{devlin-etal-2019-bert} as our PML here as it has been the focus of much prior work on bias evaluations~\cite{kaneko-bollegala-2021-debiasing,guo-etal-2022-auto,meade-etal-2022-empirical}, the evaluation protocol we use can be applied to any PLM.
We used the word lists\footnote{\url{https://github.com/kanekomasahiro/context-debias}} proposed by \newcite{kaneko-bollegala-2021-debiasing} as female words, male words, and occupational words for extracting data instances and debiasing.

We use the following nine downstream tasks from the GLEU benchmark: CoLA~\cite{warstadt-etal-2019-neural}, MNLI~\cite{williams-etal-2018-broad}, MRPC~\cite{dolan-brockett-2005-automatically}, QNLI~\cite{rajpurkar-etal-2016-squad}, QQP\footnote{\url{https://quoradata.quora.com/First-Quora-Dataset-Release-Question-Pairs}}, RTE~\cite{dagan2006pascal,haim2006second,giampiccolo2007third,bentivogli2009fifth}, SST-2~\cite{socher-etal-2013-recursive}, STS-B~\cite{cer-etal-2017-semeval}, and WNLI~\cite{levesque2012winograd}.
Hyperparameters for debiasing follow previous studies~\cite{kaneko-bollegala-2021-debiasing,webster2020measuring}, and we used the default values of huggingface for downstream task hyperparameters.\footnote{\url{https://github.com/huggingface/transformers/tree/main/examples/pytorch/text-classification}}
For fine-tuning we use the entire training dataset for each corresponding task, without splitting into male, female and occupational instances.
We evaluate the performance on all tasks using the official development data.

\subsection{Performance of Original vs. Debiased Models}

We extract instances containing female words, male words, and stereotypical words from each of the datasets.
We then calculate the performance difference between the original model and the debiased model for each dataset, and compare against the performance differences obtained when using all instances.
If the performance difference for all instances is smaller than that when evaluated for the female, male, and occupational instances, it would indicate that the effect of debiasing is underestimated when evaluated on the entire dataset.

\autoref{tbl:diff} shows the performance differences between the original model and the debiased model for each dataset/task in the GLEU benchmark.
All, Female, Male, and Occ. are the performance differences when evaluated on the entire task dataset, instances containing female words, instances containing male words, and instances containing occupational words, respectively.

From the results in \autoref{tbl:diff}, it can be seen that the performance difference between the original model and the debiased model is larger for the Female, Male, and Occ. instances compared to that when using all instances.
In particular, instances related to females exhibit a significant decrease in performance after debiasing.

It can be seen that different word lists used for debiasing have different effects on the performance degradation in downstream tasks due to debiasing.
Context debiasing uses occupational words for debiasing, while CDA debiasing does not.
Therefore, in CDA debiasing, Occ. does not have a large performance difference compared to female- and male-related instances.
Therefore, in CDA debiasing, the performance difference for occupation-related instances is smaller than that for the female and male-related instances.
On the other hand, in Context debiasing, occupation-related instances has the largest performance difference as well as female- and male-related instances.
Dropout debiasing does not use word lists for debiasing.
Therefore, unlike CDA and context debiasing, we see large drops in performance for female, male and Occ. across tasks with Dropout debiasing.

\subsection{Debias Controlled Method}

\begin{figure}[!t]
\centering

\begin{subfigure}{0.5\textwidth}
\centering
\includegraphics[width=\linewidth]{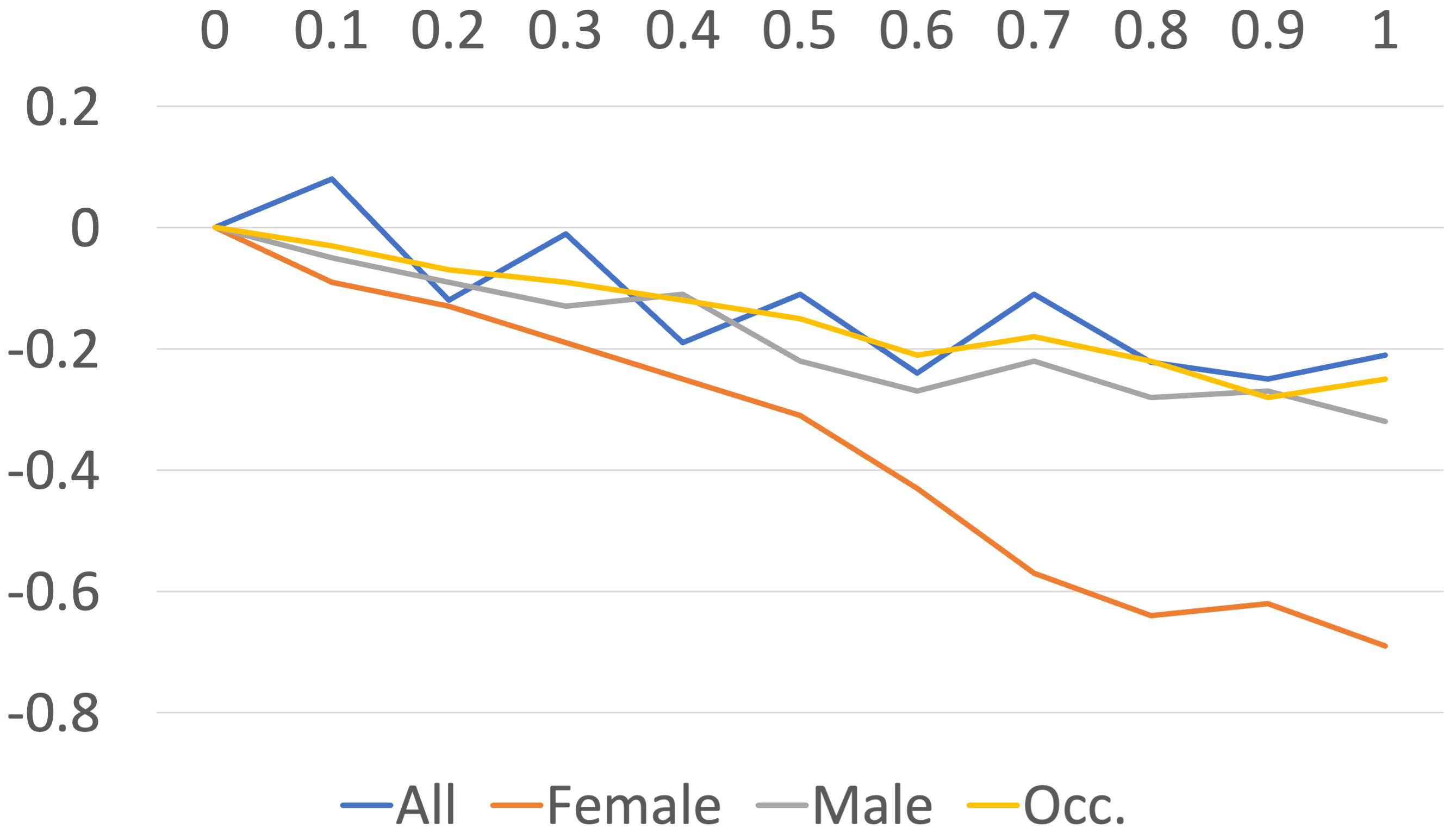}
\caption{QQP.}
\label{fig:qqp}
\end{subfigure}

\vspace{0.1cm} 

\begin{subfigure}{0.5\textwidth}
\centering
\includegraphics[width=\linewidth]{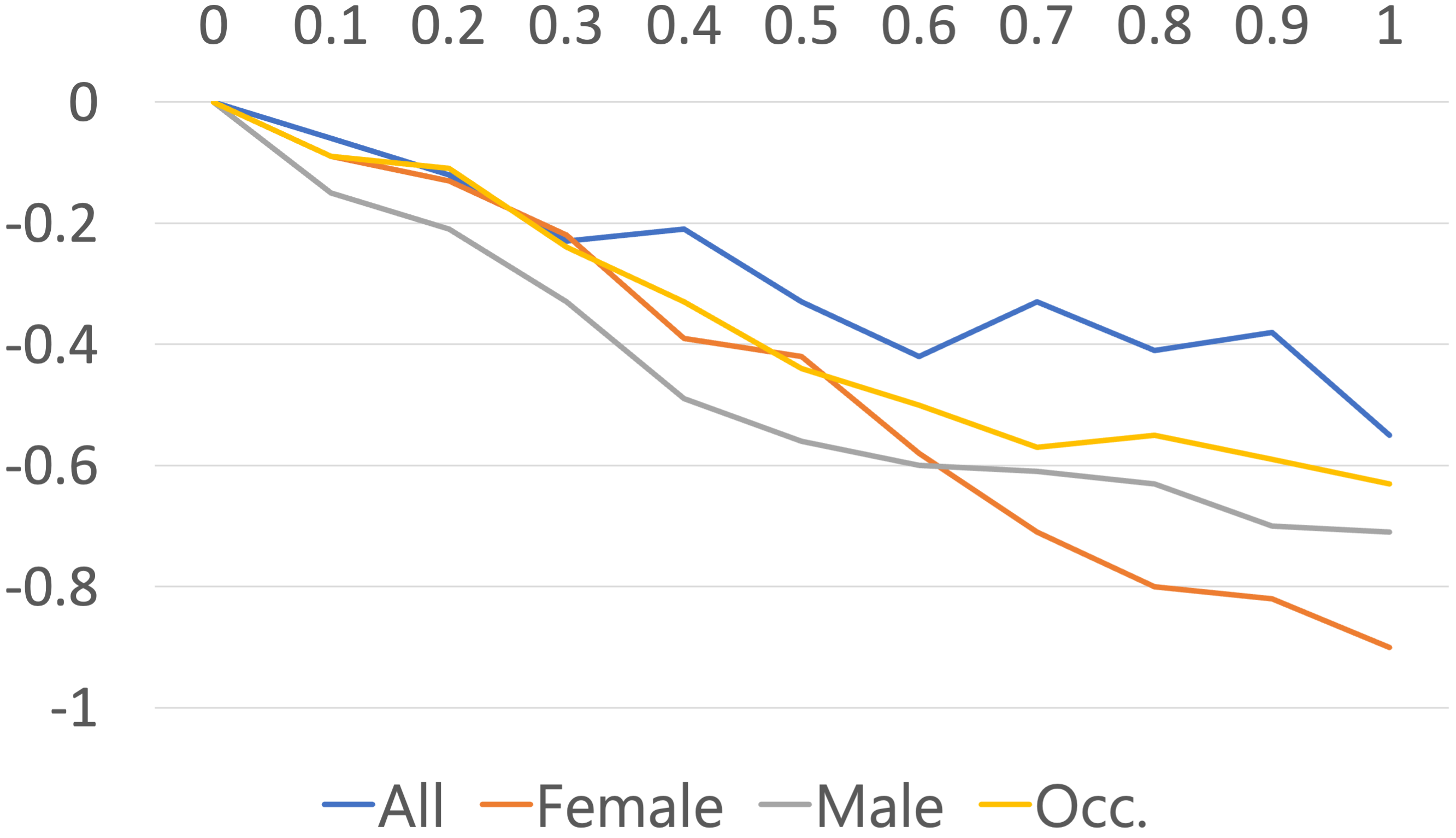}
\caption{MNLI.}
\label{fig:mnli}
\end{subfigure}

\vspace{0.1cm} 

\begin{subfigure}{0.5\textwidth}
\centering
\includegraphics[width=\linewidth]{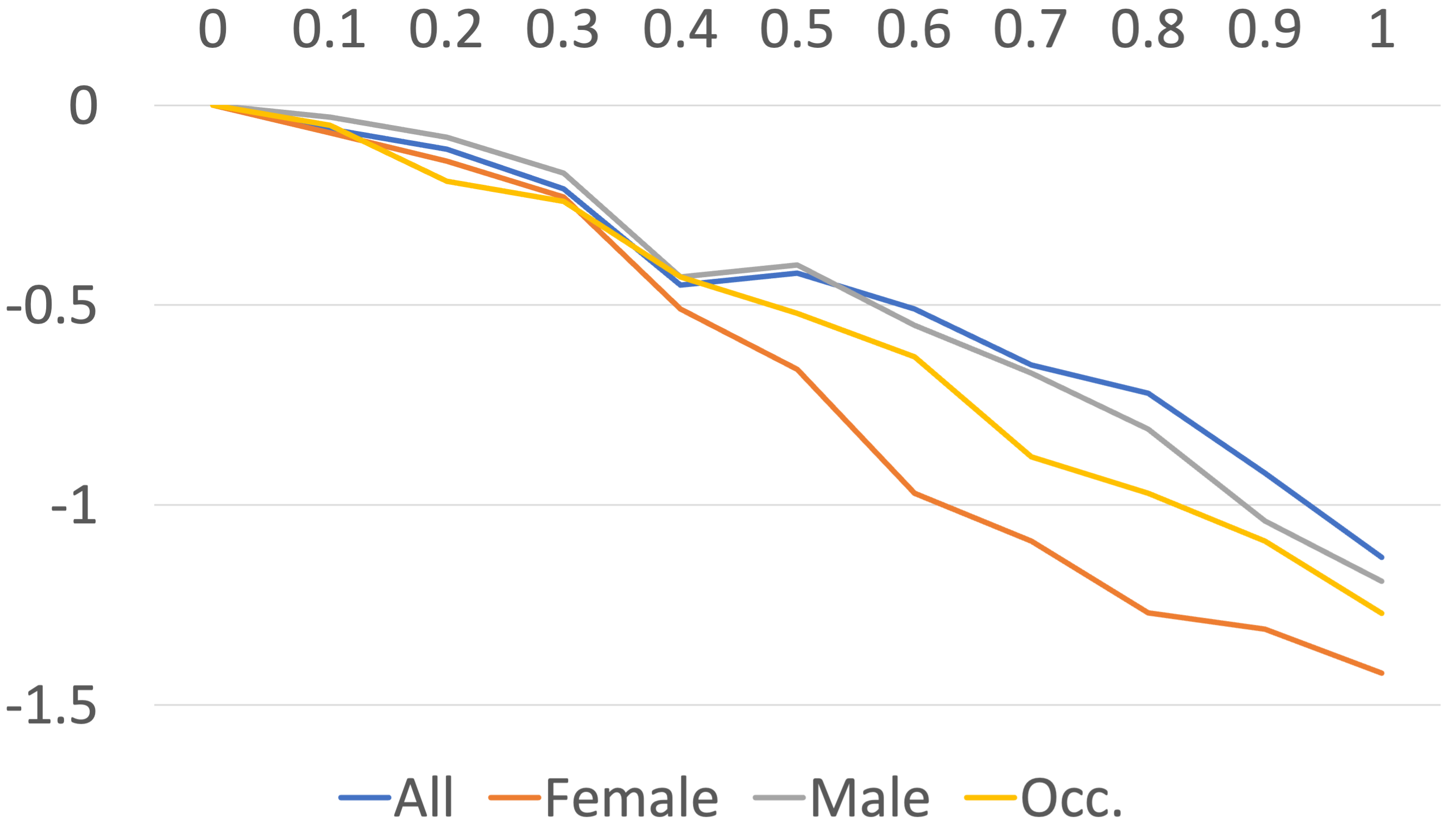}
\caption{QNLI.}
\label{fig:qnli}
\end{subfigure}

\caption{Performance difference between original and debiased models by debias rate $r$. The vertical axis shows the performance difference between the original and debiased models, and the horizontal axis shows the debias rate.}
\label{fig:debias}
\end{figure}

To understand how debiasing of an PLM affects the performance of individual downstream benchmark datasets, following the probing technique proposed by \newcite{kaneko-etal-2023-comparing}, we apply different levels of debiasing to bert-base-cased PLM and measure the difference in performance with respect to its original (non-debiased) version.
For this purpose we use CDA as the debiasing method, where we swap the gender-related pronouns in $r \in [0,1]$ fraction of the total $N$ instances of a dataset (i.e.the total number of gender swapped instances in a dataset will be $r \times N$).
$r=0$ corresponds to not swapping gender in any training instances of the dataset, whereas $r=1$ swaps the gender in all instances.
We increment $r$ in step size $0.1$ to obtain increasingly debiased version of the PLM, which is then fine-tuned for the downstream task\footnote{In \autoref{sec:appendix}, we show that the debias controlled method is able to debias the model according to $r$.}.
\autoref{fig:debias} shows the difference in performance between the original vs. debiased versions of the PLM for QQP, MNLI, and QNLI, which have the largest numbers of instances in the GLEU benchmark.

Note that CDA debiasing reverses gender without considering the context, as in \textit{``He gets pregnant''} for \textit{``She gets pregnant''}.
This is problemantic because it eliminates even useful gender-related information learnt by the PLM via co-occurring contexts.
Therefore, CDA debiasing has a negative impact on the performance of downstream tasks~\cite{zmigrod-etal-2019-counterfactual} as shown by all three subplots in \autoref{fig:debias}.
In fact, \autoref{tbl:diff} shows that the performance difference of CDA debiasing is larger than that of dropout debiasing and context debiasing.
Therefore, the larger $r$ is for CDA, the more gender instances is balanced and debiased, but the performance is unfortunately degraded.

If the dataset of the downstream task is sensitive to the effect of debiasing, the performance difference between the original model and the debiased model widens as $r$ increases.
On the other hand, if the data set is insensitive to the effect of debiasing, the performance difference between the original model and the debiased model is unlikely to increase with the value of $r$.

We find that the performance differences for the female, male, and occupational instances in the QQP, MNLI, and QNLI datasets increase with the value of $r$. 
On the other hand, for QQP and MNLI, there is a rise and fall in the performance difference when all data are used.
These results indicate that All, which includes instances related to non-gender, are less sensitive to the effect of debiasing compared to Female, Male, and Occupational instances.

On the other hand, for QNLI, All has a small rise and fall in the performance difference.
As seen from \autoref{tbl:statistics}, QNLI contains more gender-related instances than QQP and MNLI.
Therefore, it is likely that the performance decreases with $r$ even for All.
All and Male instances have a similar trend in performance difference with $r$.

\section{Conclusion}

This study focused on gender-related social biases and the presence of female, male, and stereotypical words in benchmark datasets.
Prior work had used the performance on downstream tasks to prove the usefulness of debiasing methods, overlooking the fact that only a small fraction of those downstream benchmark datasets contain gender-related instances.
On the contrary, we found that the effects of debiasing an PLM were consistently underestimated across all tasks.
We recommend that the evaluation of debiasing effects must be more reliably conducted by considering instances containing specific gender-related words separately rather than evaluating all instances in a benchmark dataset.

\section{Ethical Considerations}

This study uses existing methods and datasets for experiments and does not propose a debiasing method or create a new dataset for social bias.
This study evaluates the impact of debiasing on the performance of the downstream task, and it is not possible to evaluate how much bias is mitigated in the PLMs.
Therefore, when evaluating the bias of PLMs, it is necessary to use evaluation methods such as StereoSet~\cite{nadeem-etal-2021-stereoset}, Crowds-Pairs~\cite{nangia-etal-2020-crows}, and All Unmasked Likelihood~\cite{kaneko2022unmasking}.

In this study, we only included binary gender as a gender bias.
However, gender bias regarding non-binary gender has also been reported~\cite{cao-daume-iii-2020-toward,dev-etal-2021-harms}.
It is necessary to verify whether there is a similar trend in debiasing for non-binary genders.

\section{Limitations}

Many previous studies have shown that various social biases other than gender bias are learned in PLMs.
This study targets only gender bias.
While existing studies~\cite{webster2020measuring,zhao-etal-2019-gender} have debiasing various PLMs, we have experimented only with bert-base-cased.
Furthermore, although this study targets only English, which is a morphologically limited language.
On the other hand, various types of social biases are also learned in the PLMs across many languages~\cite{kaneko-etal-2022-gender,neveol-etal-2022-french}.
Therefore, if the proposed method is to be used with other social biases and PLMs, it is necessary to properly verify its effectiveness in languages other than English.
Moreover, we have not verified the use of debias controlled methods in languages such as Spanish and Russian, where gender swapping is not easy from a grammatical point of view~\cite{zmigrod-etal-2019-counterfactual}.

\bibliography{anthology,custom}
\bibliographystyle{acl_natbib}

\clearpage
\appendix

\section{Bias Evaluation in NLI task}
\label{sec:appendix}

We show that the debias controlled method is appropriately debiasing according to $r$.
We use  Fraction Neutra~\cite[\textbf{FN};][]{Dev_2020} as the bias evaluation method.
The FN method evaluates bias in the NLI by considering the percentage of neutral labels predicted by the model for the premise sentence (e.g. \textit{The driver owns a cabinet.}) and the hypothesis sentence (e.g. \textit{The man owns a cabinet.}) generated with the template.
The FN method indicates that the lower the score, the more bias there is in the model.
We evaluate PLMs trained on MNLI with FN method.

\begin{figure}
\centering
\includegraphics[width=0.5\textwidth]{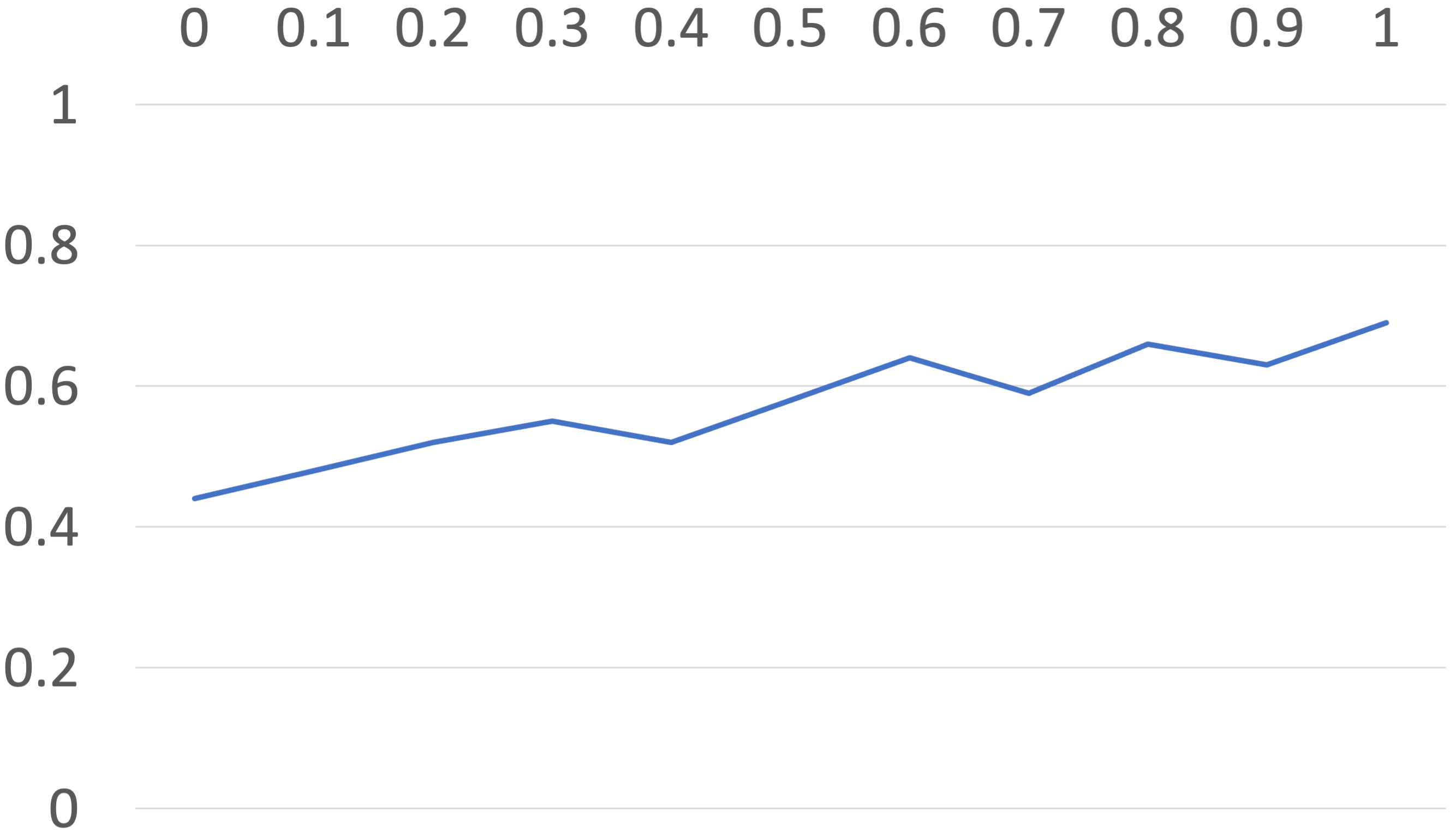}
\caption{Evaluation of debias controlled models using FN evaluation method. The vertical axis shows the bias score, and the horizontal axis shows $r$.}
\label{fig:aula}
\end{figure}

\autoref{fig:aula} shows the bias scores of FN method for each debias controlled model.
It can be seen that the bias of the model is decreasing with $r$.
Therefore, the debias controlled method is able to debias the models according to $r$.

\end{document}